\documentclass[11pt]{article}
\usepackage[utf8]{inputenc}

\usepackage{mystyle}

\begin{document}

\title {\huge Infinitely Wide Tensor Networks as Gaussian Process}

\author
{Erdong Guo\thanks{University of California, Santa Cruz, email: \texttt{eguo1@ucsc.edu}.} \qquad David Draper\thanks{University of California, Santa Cruz, email: \texttt{draper@ucsc.edu}} 
}

\date{}

\maketitle

\begin{abstract}
Gaussian Process is a non-parametric prior which can be understood as a distribution on the function space intuitively. It is known that by introducing appropriate prior to the weights of the neural networks, Gaussian Process can be obtained by taking the infinite-width limit of the Bayesian neural networks from a Bayesian perspective.
In this paper, we explore the infinitely wide Tensor Networks and show the equivalence of the infinitely wide Tensor Networks and the Gaussian Process. We study the pure Tensor Network and another two extended Tensor Network structures: Neural Kernel Tensor Network and Tensor Network hidden layer Neural Network and prove that each one will converge to the Gaussian Process as the width of each model goes to infinity. (We note here that Gaussian Process can also be obtained by taking the infinite limit of at least one of the bond dimensions $\alpha_{i}$ in the product of the chain of tensor nodes, and the proofs can be done with the same ideas in the proofs of the infinite-width cases.) We calculate the mean function (mean vector) and the covariance function (covariance matrix) of the finite dimensional distribution of the induced Gaussian Process by the infinite-width tensor network with a general set-up. We study the properties of the covariance function and derive the approximation of the covariance function when the integral in the expectation operator is intractable. In the numerical experiments, we implement the Gaussian Process corresponding to the infinite limit tensor networks and plot the sample paths of these models. We study the hyperparameters and plot the sample path families in the induced Gaussian Process by varying the standard deviations of the prior distributions. As expected, the parameters in the prior distribution namely the hyper-parameters in the induced Gaussian Process controls the characteristic lengthscales of the Gaussian Process.
\end{abstract}

%!TEX root = main.tex

\section{Introduction}
Gaussian Process (GP) as the infinite neural network function prior attracts remarkable attention recently ever since the discovery of the equivalence between \emph{GP} and the \emph{infinite-width one hidden layer Neural Network} ~\citep{neal1995bayesian}. The idea of setting the number of neurons in the hidden layer of neural network to the infinite limit is motivated by answering the following question: how our knowledge (background information) can be used to suggest the prior distribution? It turns out that as the number of neurons in the hidden layer goes to infinity, the response of the infinite-width hidden layer neural network with weights prior $p(\mathbf{w})$ converges to the GP. The interesting point of this limit is that we can obtain a prior distribution on the function space by the limiting process, namely the distribution $p(\mathbf{y})$ of the response function  by introducing a prior distribution $p(\mathbf{w})$ on the weights. More concretely, by achieving the infinite-width limit of the neural network, the response function $y(\mathbf{w}; \mathbf{x})$ is sampled from the function space with $p(y(\mathbf{w}; \mathbf{x})|\mathbf{w}, \mathbf{x})$ density function rather than parameterized by a fixed function form with the parameters to be learned from the data set.

In some sense, the GP is more flexible which has stronger representation power than the neural networks. So it is suggested to work on the non-parametric Gaussian Process model which is parameterized arbitrarily before the data arrive instead of the neural network with fixed parameterized function. D. MacKay has a nice review discussing the neural networks and Gaussian Process, see ~\citep{mackay1998introduction}. In the work by Williams ~\citep{williams1997computing}, the GPs induced by the limit of the neural network with specific neurons, namely the sigmoid and gaussian neurons, are constructed. The following work studied the infinite limit of the neural network with more general structures: the general shallow infinite-width neural network, the convolutional neural network and other structures ~\citep{pmlr-v2-leroux07a, novak2018bayesian, garriga2018deep, yang2019scaling}. In ~\citep{hazan2015steps, lee2017deep, matthews2018gaussian}, the equivalence between the infinite-width deep neural network and the GP is proved and the performance of the infinite-width deep neural network induced GP is compared with the normal model. Recently, the connection between kernel machine and sufficiently wide neural network is studied in a series work, see ~\citep{cho2009kernel, daniely2016toward, daniely2017sgd, jacot2018neural}.  

%!TEX root = main.tex

\section{Preliminaries: Gaussian Process by Neural Networks}
\subsection{Basis of Gaussian Process}
As is well known, Gaussian Process can be understood as the generalization of the multi-variate Gaussian distribution on the vector space to infinite dimensional function space which is defined by the mean function $\mu(x)$ and the covariance function $C(x, x^{\prime})$. Here $\mu(x)$ represents the mean of the random variable $X$ with index $x$, and the covariance $C(a, x^{\prime})$ represents the covariance between two random variable $X$ and $X^{\prime}$ with index $x$ and $x^{\prime}$. In the application case with finite size training data set, the finite dimensional distribution (f.d.d.s.) of the Gaussian Process is used to model the data set, and then the mean function degenerate to the finite dimensional vector $\mathbf{\mu}$ and the covariance function degenerate to the covariance matrix $C_{ij}$ which defines the f.d.d.s. of the GP, namely the multi-variates Gaussian distribution $\mathcal{N}(\mathbf{\mu}, C_{ij})$. 

The nice point of the GP is that the Bayesian updating of the GP prior is analytically tractable and simple. For a GP prior as follows, 
\begin{align}
   (f|\mu, C)\sim\mathcal{GP}(f|\mu, C).
\end{align} 
We can get the joint distribution of the training data $\mathbf{x}$ and the $\mathbf{x}^{*}$ easily as 
\begin{align}
   \begin{bmatrix}
   f\\
   f^{*}
   \end{bmatrix} 
   \sim
   \mathcal{N}(
   \begin{bmatrix}
   \mu(\mathbf{x})\\
   \mu(\mathbf{x}^{*})
   \end{bmatrix}, 
   \begin{bmatrix}
   C(\mathbf{x}, \mathbf{x}) &
   C(\mathbf{x}, \mathbf{x}^{*})\\
   C(\mathbf{x}^{*}, \mathbf{x}) & 
   C(\mathbf{x}^{*}, \mathbf{x}^{*}))
   \end{bmatrix}.
\end{align}
The predictive posterior distribution $p(f^{*}|\mathbf{x}^{*}, \mathbf{x}, f)$ can be obtained analytically as 
\begin{align}
    (f^{*}|f, \mathbf{x}^{*}, \mathbf{x})\sim\mathcal{N}(C(\mathbf{x}^{*}, \mathbf{x})C(\mathbf{x},\mathbf{x})^{-1}f, C(\mathbf{x}^{*}, \mathbf{x}^{*}) - C(\mathbf{x}^{*}, \mathbf{x})C(\mathbf{x}, \mathbf{x})^{-1}C(\mathbf{x}, \mathbf{x}^{*}).
\end{align}
It is easy to extend above analysis to the noisy label model, namely introducing noise to the label $\mathbf{y} = f(\mathbf{x}) + \mathcal{\epsilon}$. 

For a non-parametric model, it can adapt itself to simulate the data after the data comes but we can still use the Maximum Likelihood Estimator (M.L.E.) or the Maximum A Posterior (M.A.P.) to optimize the hyper-parameters. We can get the nice marginal likelihood as follows,
\begin{align}
    \log{(\mathbf{y}|\mathbf{x}, \mathbf{\theta})} \propto -\frac{1}{2}\mathbf{y}^{T}C_{\mathbf{y}}^{-1}\mathbf{y}-\frac{1}{2}\log{|C_{\mathbf{y}}|},
\end{align}
where $C_{\mathbf{y}} = C_{f} + \sigma_{n}^{2}\mathbb{I}$. 

The training data only get into first term of above formula which evaluates how good our model fits the given training data, however the second term only depends on the covariance matrix which is a model complexity penalty term. We can undertand the competition between the data fitting and the model capacity by considering the length-scale. The model capacity will increase as the length-scale decrease, namely the model complexity penalty will increase although the data-fitting will get better. From the computation perspective, we know to get the optimal point of the marginal likelihood, we need to compute the inverse of the covariance matrix $C_{\mathbf{y}}^{-1}$ which is of order $O(n^{3})$, where $n$ is the training sample size. For application in the big data set, the inverse matrix computation will be highly time-consuming so we can consider using matrix approximation method such as outer-product approximation.
See ~\citep{williams2006gaussian} for a good review on GP application in Machine Learning.

For an infinite-width neural network with response function $f(\mathbf{x})$ as 
\begin{align}
    f(\mathbf{x}) = b^{[1]} + \sum_{i}w^{[1]}_{i}a_{i}(\mathbf{w}^{[0]}; \mathbf{x}),
\end{align}
where $\mathbf{w}^{[i]}$ and $b^{[i]}$ represent the weights and the bias in the $i$'th layer and $a_{i}(\mathbf{w}; \mathbf{x})$ represents the activation function. Since all parameters in the neural network are independent and identically distributed (i.i.d.), by the Central Limit Theorem (C.L.T.), we know as $i\to\infty$,
\begin{align}
\sum_{i}w_{i}^{[1]}a_{i}(\mathbf{w}^{[0]}; \mathbf{x}) \to X,
\end{align}
where $X$ belongs to the Normal distribution. Since $b^{[1]}$ is normal and independent with $\sum_{i}w_{i}^{[1]}a(w_{i}^{[0]}; x)$, we know $f(\mathbf{x})$ is normal random variable. We will get a sequence of random variables $\{f(\mathbf{x}^{(i)}), i\in\{1,\cdots, m\}\}$ by evaluating the infinite neural network on the data set.
For the neural network with several special non-linear activation functions, analytical expressions of the Gaussian Process are obtained ~\citep{williams1997computing}.

Above discussion is an intuition explaining of the equivalence between GP and infinite-width neural networks, here we will propose our first theorem on the equivalence between GP and neural networks and provide a strict proof in the Appendix A. 
\begin{theorem}
\label{the: neural_net_eq_gp}
Let the response $f(\mathbf{x})$ of the one hidden layer neural network with $j$ neurons as 
\begin{align}
    f(\mathbf{x}) = b^{[1]} + \sum_{j}w^{[1]}_{j}a_{j}(\mathbf{w}^{[0]}; \mathbf{x}),
\end{align}
where
\begin{align*}
   &w_{j}^{[i]} \overset{\text{i.i.d.}}{\sim} \mathcal{N}(0, \sigma_{i}),\\
   &b^{[i]} \sim \mathcal{N}(0, \sigma_{b}).
\end{align*}
As $j$ goes to infinity, $f$ converge to a Gaussian Process,
%\begin{align*}
%    f\sim \mathcal{GP}(\mu, \Sigma),
%\end{align*}
$\mathcal{GP}(\mu, \Sigma)$ with $\mu$, and $\Sigma$ as the mean and covariance function. 
\end{theorem}

%!TEX root = main.tex

\section{Infinite Limits of Tensor Networks}
\subsection{Introduction of Tensor Network}
For a $n$ nodes Tensor Network, actually Matrix Product States (M.P.S.) ~\citep{biamonte2017tensor, orus2014practical, cichocki2014era}, the scalar response $f(\mathbf{x})$ is as follows,
\begin{align}
    \psi(\mathbf{x}) = \sum_{\{\alpha_{i}, s_{i}\}}A^{s_{1}}_{\alpha_{1}\alpha_{2}}\cdots A^{s_{i}}_{\alpha_{i}\alpha_{i+1}}\cdots A^{s_{n}}_{\alpha_{n}\alpha_{1}}\Phi^{s_{1}\cdots s_{n}}(\mathbf{x}),
\end{align}
where
\begin{align*}
    \Phi^{s_{1}s_{2}\cdots s_{n}}(\mathbf{x}) = \phi^{s_{1}}(x_{1})\otimes\cdots\phi^{s_{i}}(x_{i})\cdots\otimes\phi^{s_{n}}(x_{n}).
\end{align*}
Here $\phi^{s_{i}}(x_{i})$ represents the function which maps each component of the input data $x_{i}$ into a higher dimensional feature space. $A^{s_{i}}_{\alpha_{i}\alpha_{i+1}}$ represents the rank $3$ tensor node, where $s_{i}$ is the physical index which contracts with the $s_{i}$ index in the kernel $\phi^{s_{i}}$. $\alpha_{i}$ is the bond index by which we can tune the representation power of the tensor network.

M.P.S. is introduced to approximate a quantum state $\Ket{\psi}$ efficiently in quantum physics ~\citep{orus2019tensor, chabuda2020tensor, schroder2019tensor, mcmahon2020holographic}. To fully describe a $n$-party quantum state $\Ket{\psi}$, we need $2^{n}$ basis which grows exponentially with respecting to $n$. By expanding the $n$-party quantum state with a chain of product of matrices, we can approximate the original quantum state with less parameters. For the M.P.S., the number of the basis grows linearly as $n$ increase. 
\begin{align}
    \Ket{\psi} = \psi_{s_{1}\cdots s_{n}}\Ket{s_{1}\cdots s_{n}}\approx A^{s_{1}}_{\alpha_{1}\alpha_{2}}\cdots A^{s_{n}}_{\alpha_{n}\alpha_{1}}\Ket{s_{1}\cdots s_{n}}
\end{align}
Here we write a general n-party state as a matrix product chain with $O(s\alpha^{2}n)$ parameters.

Tensor Networks as a powerful machine has been widely studied to construct statistical learning model in Machine Learning community recently and also great achievements have been obtained in different tasks such as supervised pattern recognition (classification and regression), natural language processing and density estimation (unsupervised generative model) ~\citep{stoudenmire2016supervised, han2018unsupervised, pestun2017language, novikov2018exponential}. Since the special structure of tensor network, the model easily blows out or decays to dead nodes during training process, efficient and robust initialization strategy was proposed in ~\citep{2101.00245}. It is known that the correlation function of tensor networks decays exponentially ~\citep{evenbly2011tensor, evenbly176quantum}, so it is discussed to extend the tensor network to more complicated structures such as multi-scale tree structure to reduce the correlation decay problem in machine learning application ~\citep{stoudenmire2018learning}. More extension work of tensor network can be found in ~\citep{cheng2020supervised, li2018shortcut, cheng2020supervised, liu2018machine, glasser2018supervised}.

\subsection{Infinite-Width Pure Tensor Networks}
\subsubsection{Theory}
Since the special structure of the MPS which is the contraction of a chain of tensor nodes, we can obtain a 'trivial' Gaussian Process pure MPS without activation. So our first theorem on the infinite-width MPS equivalent GP is as follows,
\begin{theorem}
\label{the: pure_mps_eq_gp}
For an infinite-width MPS with independent random tensor nodes (we note here i.i.d. is not necessary), namely
\begin{align}
   &(A^{s_{i}}_{\alpha_{i} \alpha_{i+1}}|\mathbf{\theta}_{i})\sim p(A^{s_{i}}_{\alpha_{i}\alpha_{i+1}}|\mathbf{\theta}_{i}), 
\end{align}
with at lease one of the bond index whose dimension $|a_{i}| \geq 2$, 
as the number of the tensor nodes goes to infinity, the response will converge to the Gaussian Process,
\begin{align}
    \psi\sim\mathcal{GP}(\mu, \Sigma),
\end{align}
where $\psi$ is the response of the infinite-width MPS, $\mu$ is the mean function and $\Sigma$ is the covariance function. 

\noindent
Note: the GP limit can also be obtained by taking the infinite limit of at lease one of bond index $\alpha_{i}$, where $i\in\{1, \cdots, n\}$ if the width of the tensor network is fixed to be finite.
\end{theorem}

A proof of Theorem \ref{the: pure_mps_eq_gp} is provided in Appendix \ref{app: pure_mps_eq_gp_proof}. 
%\begin{theorem}
%For an infinite width tensor network with independent and partly identical prior distribution and deterministic kernel, as the number of the tensor nodes goes to infinity, the response distribution will converge to the Gaussian Process.
%\end{theorem}
Without hidden layer and activation, the response of MPS is linear with respect to the tensor weights $A^{s_{i}}_{\alpha_{i}, \alpha_{i+1}}$ which means the sample path of the equivalent GP is linear function. We can write down the joint distribution of the sequence of random variables, namely the response of the MPS evaluating at different training points $\psi(\mathbf{x}^{(i)})$. 
\begin{align}
    p(\psi_{1}, \cdots, \psi_{m}) = p(\psi_{1})\delta{(\psi_{1}-k_{2}\psi_{2}})\cdots\delta{(\psi_{1}-k_{m}\psi_{m}}),
\end{align}
where $\psi_{i} = \psi(x^{(i)})$, $k_{i}$ is the coefficient of the linear transformation between $\psi_{i}$ and $\psi_{1}$, namely $\psi_{i} = k_{i}\psi_{1}$ and $p(\psi_{1})$ is the normal distribution. It is easy to verify that each random variable $\psi_{i}$ is normal  and then the sequence $\{\psi_{i}, i\in\{1, \cdots, m\}\}$ is GP. 

Formally we can write down the joint distribution of $\psi_{i}$ and $\psi_{j}$ as
\begin{align*}
    p(\psi_{i}, \psi_{j}|A^{s_{i}}_{\alpha_{i}\alpha_{j}}) = \mathcal{N}(\psi_{i}, \psi_{j}|\mathbf{0}, \Sigma),
\end{align*}
where 
\begin{align*}
    \Sigma = 
\begin{bmatrix}
\sigma_{1}^{2} & \rho\sigma_{1}\sigma_{2}\\
 \rho\sigma_{1}\sigma_{2} & \sigma_{2}^{2}
\end{bmatrix}.
\end{align*} 
Since any two random variables $\psi_{i}$ and $\psi_{j}$ in the sequence are related by a linear transformation, we know the correlation $\rho$ of these two random variables is $1$ or $-1$.
Using the condition probability rule, we have 
\begin{align*}
    p(\psi_{i}, \psi_{j}|A^{s_{i}}_{\alpha_{i}\alpha_{j}}) &= p(\psi_{i}|A^{s_{i}}_{\alpha_{i}\alpha_{j}})p(\psi_{j}|\psi_{i}, A^{s_{i}}_{\alpha_{i}\alpha_{j}}),
\end{align*}
where
\begin{align*}
p(\psi_{j}|\psi_{i}, A^{s_{i}}_{\alpha_{i}\alpha_{j}}) &= \lim_{\sigma\to 0}\mathcal{N}(\psi_{j}-\rho\frac{\sigma_{j}}{\sigma_{i}}\psi_{i}, \sigma)\\
&=\delta{(\psi_{j}-\rho\frac{\sigma_{j}}{\sigma_{i}}\psi_{i})}.
\end{align*}
So in the pure MPS case, the GP is 'trivial' since the uncertainty band of $\psi_{j}$ given $\psi_{i}$ is $0$, and then the conditional distribution $p(\psi_{j}|\psi_{i},A^{s_{i}}_{\alpha_{i}\alpha_{j}})$ is a generalized function which is defined as the limit of a sequence of normal distributions showed above.

We can get the sample path of the GP as follows,
\begin{align*}
   \psi(x) = \frac{\psi_{2}-\psi_{1}}{\psi_{2}-\psi_{1}}(x - x_{1}) + \psi_{1},
\end{align*}
where $\psi_{1}$ and $\psi_{2}$ are two responses of the MPS evaluating at data points $x^{(1)}$ and $x^{(2)}$.

\subsubsection{Discussion on Equivalent GP induced by Specific Kernel}
%For a specific prior distribution as
%\begin{align}
%    p(A^{s_{i}}_{\alpha_{i}\alpha_{i+1}}|\mathbf{\theta}_{i}) = \mathcal{N}(A^{s_{i}}_{\alpha_{i}\alpha_{i+1}}|0, \sigma_{i}^{2})
%\end{align}
We can get the mean function $\mu(\mathbf{x})$ and the covariance function $\Sigma(\mathbf{x}, \mathbf{x}^{\prime})$ of the equivalent GP of pure tensor network as follows,
\begin{align}
   &\mu(x^{(i)}) = \mathrm{E}[\psi(x^{(i)})] = 0,\\
   &\Sigma(x^{(i)}, x^{(j)}) = \mathrm{E}[\psi(x^{(i)})\psi(x^{(j)})] = \alpha^{n-1}\prod_{k}\mathrm{E}_{A}[Z(A; x^{(i)}_{k})Z(A; x^{(j)}_{k})]
\end{align}
where
\begin{align*}
   Z(x_{i}) = 
   A^{s}_{\alpha_{i-1}\alpha_{i}}\phi^{s}(x_{i}).
\end{align*}
Here we consider a specific normal prior and kernel as follows,
\begin{align*}
    &p(A^{s}_{\alpha_{i}\alpha_{i+1}}|\mathbf{\sigma_{A}})\sim\mathcal{N}(A^{s}_{\alpha_{i}\alpha_{i+1}}|0, \sigma_{A})\\
    &\phi(x_{i}) = [f(x_{i}), g(x_{i})] = [x_{i}, 1-x_{i}],
\end{align*}   
we get the mean function and the covariance function as 
\begin{align*}
&E_{A}[\psi(\mathbf{x}; A)^{2}] = \alpha^{n-1}\sigma_{A}^{2n}\prod_{i}(2x_{i}^{2}-2x_{i}+1),\\
&E_{A}[\psi(\mathbf{x}; A)\psi(\mathbf{x}^{\prime}; A)] = \alpha^{n-1}\sigma_{A}^{2n}\prod_{i}(2x_{i}x^{\prime}_{i}-x_{i}-x_{i}^{\prime}+1).
\end{align*}
To avoid the power explosion problem, in application case we need to set that 
\begin{align*}
    \alpha^{n-1}\sigma_{A}^{2n}  = 1, 
\end{align*}
namely,
\begin{align*}
   \sigma_{A} = \alpha^{\frac{1-n}{2n}}.
\end{align*}

%\begin{figure}[!h]
%\begin{subfigure}{.5\textwidth}
%\centering
%\includegraphics[width=0.9\linewidth]{imgs/one_layer_mps.eps}
%\caption{plot of randomly sampled $4$ sample paths of pure MPS.}
%\label{sample_path_mps}
%\end{subfigure}
%\begin{subfigure}{.5\textwidth}
%\centering
%\includegraphics[width=0.9\linewidth]{imgs/mps_var_comparing.eps}
%\caption{plot of four sample path from the models with different variances.}
%\label{sample_path_mps_vars}
%\end{subfigure}
%\caption{In above figure, we plot the sample paths of the MPS model. Since no actvations are used in this model, so all the paths are just linear functions. Here we sampled four sample paths from the mps model. The structure of our model is as follows: the number of tensor nodes is $5$, the bond dimension is $8$ and the kernel dimension is $2$.}
%\label{fig: pure_mps_sample_path}
%\end{figure}

In the infinite limit, we should set $\sigma_{A}$ to be $\alpha^{-\frac{1}{2}}$. Actually this means in the case when the tensor network is very wide, we need to fine tune the standard deviation of the prior distribution to keep the network in stable range which is discussed in ~\citep{2101.00245}. In Fig. \ref{fig: pure_mps_sample_path}, we show the sample path of the pure MPS and also plot the sample paths sampled from the MPS with different standard deviation $\sigma$.

\subsection{Hybrid Tensor Neural Networks}
For the infinite-width pure MPS, no activation functions and no latent variables are introduced and then the response function $\psi(\mathbf{x})$ is linear respecting to the weights $A^{\alpha_{i}}_{\alpha_{i}\alpha_{i+1}}$ in each tensor node. The representation power of the pure tensor network is limited since the pure MPS can only represent the linear function. To increase the representation capacity of MPS, we can introduce another layer of neural network with activation functions. Here we propose two methods to combine the MPS with the neural network:
\begin{itemize}
    \item  Neural network kernel MPS;
    \item  MPS hidden layer neural network.
\end{itemize}
For the neural network kernel MPS, we use one layer neural network as the kernel function and 'non-linearity' is introduced by the hidden layer and the activation function in neural network. For the MPS hidden layer neural network, the response  $\psi^{l}(\mathbf{w},\mathbf{A}; \mathbf{x})$ of tensor network is fed into the neural network by which the linearity is injected into the model.

In the following sections, we will discuss these two hybrid tensor neural network and the theorems on their Gaussian Process equivalence properties.

\subsubsection{Infinite-width Neural kernel Tensor Networks}
The set-up of neural network kernel MPS is as follows,
\begin{align}
    &a_{i}(\mathbf{w}^{[0]}; \mathbf{x}) = a(\sum_{j}w^{[0]}_{ij}x_{j}), \\
    %&f^{l}(\mathbf{w}^{[1]};\mathbf{x}) = b^{l}+\sum_{k}w^{[1]}_{lk}a_{k}\\
    &\psi(\mathbf{A}, \mathbf{w}; \mathbf{x}) = \sum_{\{s, \alpha\}}A^{s_{1}}_{\alpha_{1}\alpha_{2}}\cdots A^{s_{i}}_{\alpha_{i}\alpha_{i+1}}\cdots A^{s_{n}}_{\alpha_{n}, \alpha_{1}}\Phi^{s_{1}\cdots s_{n}}(\mathbf{w}; \mathbf{x}),
\end{align}
where 
\begin{align*}
    &\phi^{s_{i}}(\mathbf{w}; \mathbf{x}) = [a_{(i-1)s+1}, \cdots, a_{is}],\\
    &\Phi^{s_{1}\cdots s_{n}}(\mathbf{w}; \mathbf{x})=\phi^{s_{1}}\otimes\cdots\otimes\phi^{s_{n}}=[[a_{1}, \cdots, a_{s}], \cdots, [a_{(n-1)s+1}, a_{ns}]],
\end{align*}
where $\mathbf{w}$ is the weights of the neural kernel, $f(\cdot)$ is the output of the hidden neural network. After we get the response of the neural network, we reshape the output of our neural network to $(n, s)$ tensor and feed it into the MPS. 

The following theorem states the equivalence between the infinite-width neural kernel tensor network and GP.
\begin{theorem}
\label{the: neural_kernel_mps_eq_gp}
For a neural kernel tensor network with independent (we note here that i.i.d. is not necessary) tensor nodes whose distribution are
\begin{align*}
   (A^{s_{i}}_{\alpha_{i} \alpha_{i+1}}|\mathbf{\theta}_{i})\sim p(A^{s_{i}}_{\alpha_{i}\alpha_{i+1}}|\mathbf{\theta}_{i}),
\end{align*}
and also with neural weights $\mathbf{w}$ whose components are independent following the distribution as 
\begin{align*}
    (w_{ij}|\sigma_{w}) \sim p(w_{ij}|\sigma_{w}),
\end{align*}
with at lease one of the bond index whose dimension $|a_{i}| \geq 2$.
As the number of the tensor nodes and also the number of the nodes of the hidden layer and the output layer of the neural network go to infinity, the response of the neural kernel tensor network will converge to a GP,
\begin{align*}
    \psi\sim\mathcal{GP}(\mu, \Sigma),
\end{align*}
where $\psi$ is the response of the infinite-width MPS, $\mu$ is the mean function and $\Sigma$ is the covariance function. 

\noindent
Note: the GP limit can also be obtained by taking the infinite limit of at lease one of bond index $\alpha_{i}$, where $i\in\{1, \cdots, n\}$ if the width of the tensor network is fixed to be finite.
\end{theorem}

A proof of above Theorem \ref{the: neural_kernel_mps_eq_gp} is given in Appendix \ref{app: neural_kernel_mps_eq_gp_proof}. The interesting point of above theorem is that we just need to assume that the tensor nodes in tensor network $A^{s_{i}}_{\alpha_{i}\alpha_{i+1}}$ to be independent with each other and also all the components of neural weights to be independent, then the respond of the model will converge to GP without assuming any other condition such as identical distribution requirements in the infinite-width neural network case. Roughly, the reason why the identical distribution requirement is not needed in our case is the factorization property of the tensor network which decompose a high rank tensor into a series of production of low rank tensor nodes. Thanks for the bond index $\alpha_{i}$ in the tensor network, the sum of the infinite i.i.d. random variables is obtained and by C.L.T, we will get a sequence of normal random variables.

The mean function $\mu{(\mathbf{x})}$ and covariance function $\Sigma(\mathbf{x}, \mathbf{x}^{\prime})$ of the neural kernel tensor network is as follows, 
\begin{align*}
   &\mathrm{E}_{A, \mathbf{w}}[\psi(x; A, \mathbf{w})] = 0,\\
   &\mathrm{E}_{A, \mathbf{w}}[\psi(x; A, \mathbf{w})\psi(x^{\prime}; A, \mathbf{w})] = 2^{n}\alpha^{n-1}\sigma_{A}^{2n}\mathrm{E}_{\mathbf{w}}^{n}[a(x; \mathbf{w})a(x^{\prime};\mathbf{w})],\\
   &\mathrm{E}_{A, \mathbf{w}}[\psi(x; A, \mathbf{w})\psi(x; A, \mathbf{w})] = 2^{n}\alpha^{n-1}\sigma_{A}^{2n}\mathrm{E}_{\mathbf{w}}^{n}[a(x; \mathbf{w})^{2}].
\end{align*}

When the activation function $a(x; \mathbf{w})$ is the Sigmoidal or Gaussian function, analytical results of $\mathrm{E}_{\mathbf{w}}[a(x; \mathbf{w})a(x^{\prime}; \mathbf{w})]$ are obtained in \cite{williams1997computing}.
In Fig. \ref{fig: neural_kernel_mps}, we plot the sample path of the MPS with neural kernel.

\subsection{Infinite-Width MPS hidden layer Neural Networks}
\subsubsection{Theory}
In this part, we feed the output of the MPS with an activation into a fully connected neuron layer. We write down the formula of the $l$ dimensional one hidden layer MPS Neural Network as follows, 
\begin{align}
    &\psi^{l}(\mathbf{x}) = \sum_{\{\alpha_{i}, s_{i}\}}A^{s_{1}}_{\alpha_{1}\alpha_{2}}\cdots A^{s_{i}l}_{\alpha_{i}\alpha_{i+1}}\cdots A^{s_{n}}_{\alpha_{n}\alpha_{1}}\Phi^{s_{1}\cdots s_{n}}(\mathbf{x}),\\
    &f(\mathbf{x}) = b + \sum_{l}w_{l}a(\psi^{l}).
\end{align}
where $a(\cdot)$ is the activation function, usually it is the Sigmoid, Relu or Tanh function.

In the following part, we will state the theorem on the equivalence between the infinite-width MPS hidden layer Neural Network and the GP. 

\begin{theorem}\label{the: mps_hidden_neural_net_eq_gp}
For the MPS hidden layer Neural Network whose tensor nodes $A^{s_{i}}_{\alpha_{i}\alpha_{i+1}}$ are all independent with each other, and also all the components $w_{ij}$ of the neural weights $\mathbf{w}$ are all independent with each other with following distributions, 
\begin{align}
   &A^{s_{i}}_{\alpha_{i} \alpha_{i+1}}\sim p(A^{s_{i}}_{\alpha_{i}\alpha_{i+1}}|\mathbf{\theta}_{i}),\\
   &(w_{ij}|\sigma_{w}) \sim p(w_{ij}|\sigma_{w}),
\end{align} 
with at lease one of the bond index whose dimension $|a_{i}| \geq 2$.
As the number of the tensor nodes goes to to infinity, the response will converge to a GP,
\begin{align*}
    f\sim\mathcal{GP}(\mu, \Sigma),
\end{align*}
where $\psi$ is the response of the infinite-width MPS, $\mu$ is the mean function and $\Sigma$ is the covariance function.

\noindent
Note: the GP limit can also be obtained by taking the infinite limit of at lease one of bond index $\alpha_{i}$, where $i\in\{1, \cdots, n\}$ if the width of the tensor network is fixed to be finite.
\end{theorem}

A proof of Theorem \ref{the: mps_hidden_neural_net_eq_gp} will be given in the appendix \ref{app: mps_hidden_neural_net_eq_gp_proof}. We note here as the width of the tensor network gets to infinity, each component of the respond of the tensor network will converge to a GP and they are independent with each other. The number of neurons in the neural network will get to infinity as the infinite-width of the tensor network is obtained, so C.L.T. can be used to show that the sequence of the random variables with the data point indices is a GP.

We calculate the mean function and the variance function as 
%\begin{align*}
%    &A^{s_{i}}_{\alpha_{i}\alpha_{i+1}}\sim \mathcal{N}(A^{s_{i}}_{\alpha_{i}\alpha_{i+1}}|0, \sigma^{2}_{A}).\\
%    &b\sim \mathcal{N}(b|0, \sigma_{b}^{2}),\\
%    &w_{l}\sim\mathcal{N}(w_{l}|0, \sigma_{w}^{2}),
%\end{align*}
\begin{align}
    &\mathrm{E}_{A, w, b}[f(\mathbf{x})] = \mathrm{E}[b] + \sum_{l}\mathrm{E}[w_{l}]\mathrm{E}[a(\psi^{l})]= 0,\\
    &\mathrm{E}_{A}[f(\mathbf{x})f(\mathbf{x}^{\prime})] = 
    \sigma_{b}^{2} + \sigma_{w}^{2}
    \mathrm{E}_{A}[\sum_{l}a^{l}(\mathbf{x})a^{l}(\mathbf{x}^{\prime})].
\end{align}
In computation of the covariance matrix, the neural activations of different data points are coupled with each other which means the analytical result is difficult to be obtained. We can use the Monte Carlo simulation to compute the matrix element of the covariance function evaluated at different data pairs by sampling hierarchically. We can get the sample activations from neural network by sampling from the prior and then plugging the prior sample into the hybrid MPS neural network which is straightforward. If the tensor network structure is not huge, the time complexity is acceptable.

In Fig. \ref{fig: mps_hidden_neural_network}, we plot the sample path of the MPS hidden layer neural network. In Fig. \ref{fig: 2d_mps_hidden_nerual_network}, we plot the two dimensional sample path of the MPS hidden layer Neural Network.

\subsubsection{Approximation of Covariance matrix}
For one hidden layer neural network, when the activation is chosen as the Sigmoidal or Gaussian function, analytical result of the covariance function was obtained ~\citep{williams1997computing}. However, in our MPS hidden neural network case, the covariance integral is much more complicated since different component of the MPS couples with each other by the bond dimension of the tensor network. We obtain the approximation of $\mathrm{E}_{\mathbf{A}}[a^{l}(\mathbf{x})a^{l}(\mathbf{x}^{\prime})]$ by expanding the square terms around the mean using Taylor expansion.
\begin{align*}
 \mathrm{E}_{A}[a^{l}(\mathbf{x};A)a^{l}(\mathbf{x}^{\prime}; A)] 
    &\approx a^{l}(\mathbf{x}; A_{0})a^{l}(\mathbf{x}^{\prime}; A_{0}) + \frac{1}{2}\mathrm{Var}[A]\cdot(2\nabla a^{l}(\mathbf{x};A_{0})\otimes\nabla a^{l}(\mathbf{x}^{\prime}; A_{0})\\
    &+\nabla^{2}a^{l}(\mathbf{x};A_{0})\otimes a^{l}(\mathbf{x}^{\prime}; A_{0})+ a^{l}(\mathbf{x};A_{0})\otimes\nabla^{2} a^{l}(\mathbf{x}^{\prime}; A_{0})),    
\end{align*}
where $\mathrm{Var}[A]$ is the diagonal part of the covariance matrix of random vector $A$, namely the variance vector every component of which is the variance of each component of the tensor nodes vector $A$. See the Appendix \ref{app: cov_matrix_approx} to get more details of the calculation.

%!TEX root = main.tex

\section{Experiments}
We implement all three tensor network induced GPs in the numerical experiments. We study the sample paths of each GP and also explore the sample paths family by varying hyperparameters, namely the standard deviations $\sigma_{\mathbf{A}}$ and $\sigma_{\mathbf{w}}$ of the tensor nodes and the neural weights prior.

\begin{figure}[!ht]
\begin{subfigure}{.49\textwidth}
\centering
\includegraphics[width=0.9\linewidth]{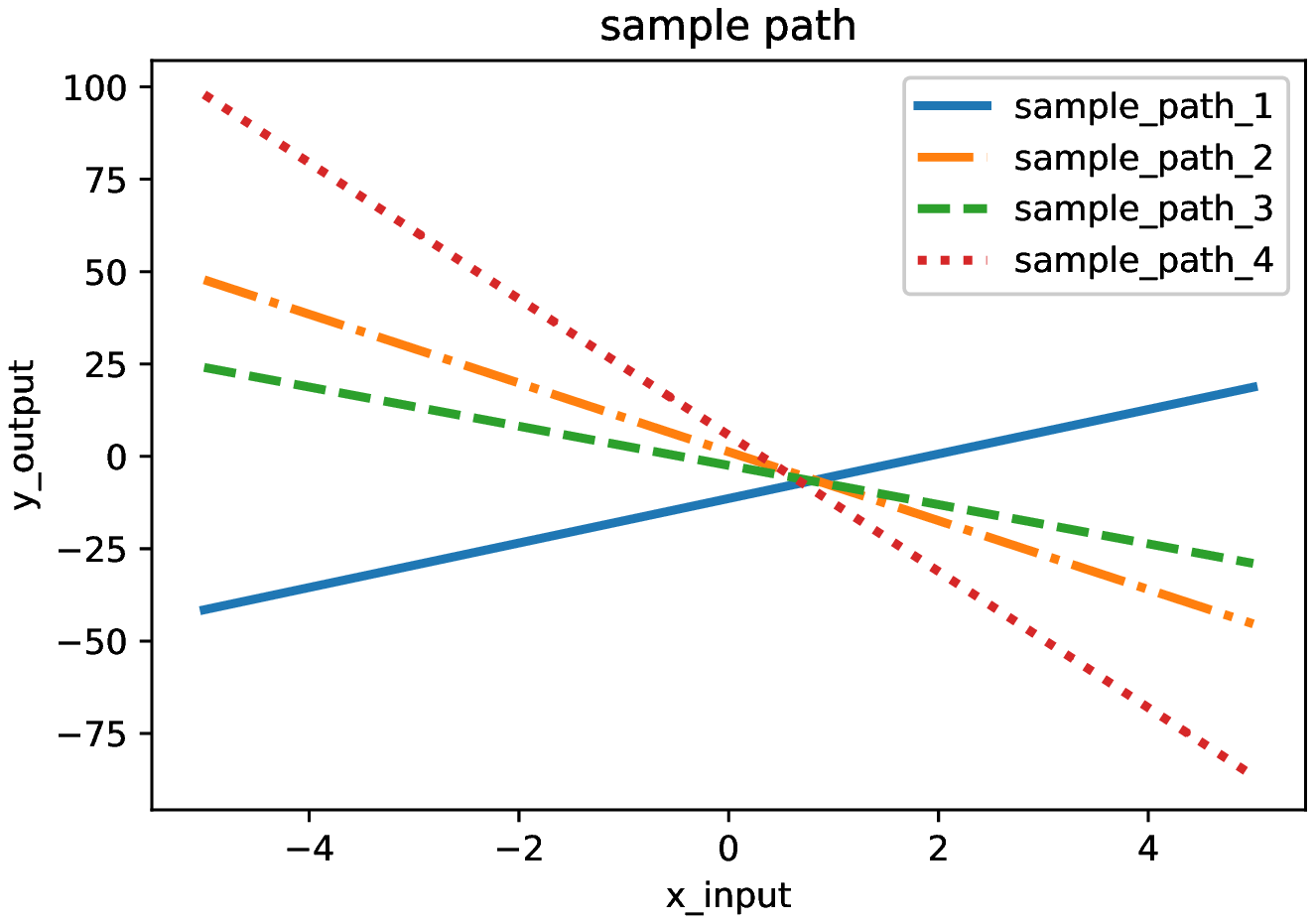}
\caption{plot of randomly sampled $4$ sample paths of pure MPS.}
\label{fig: sample_path_mps}
\end{subfigure}
\begin{subfigure}{.5\textwidth}
\centering
\includegraphics[width=0.9\linewidth]{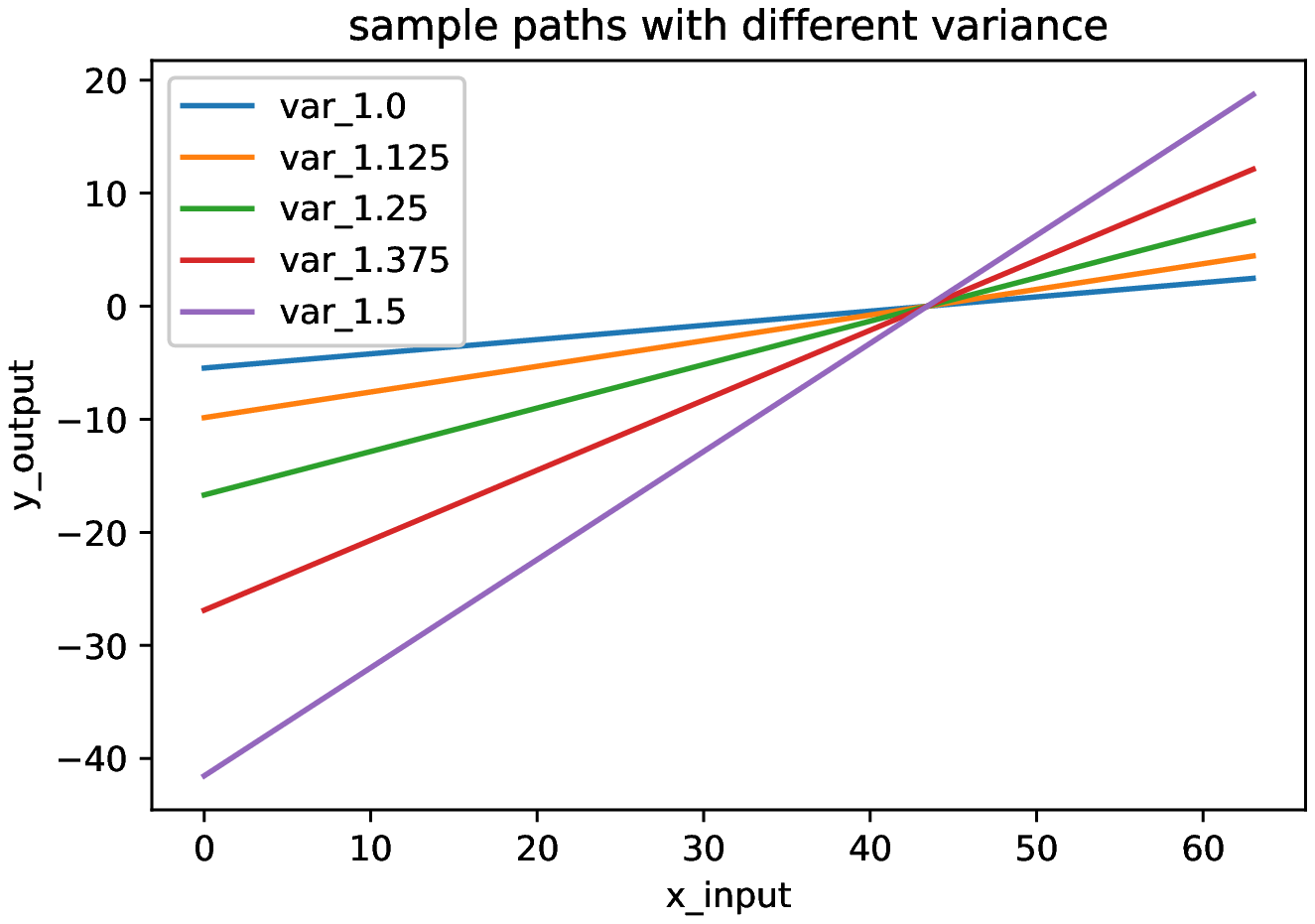}
\caption{plot of four sample path from the models with different variances.}
\label{fig: sample_path_mps_vars}
\end{subfigure}
\caption{In above figure, we plot the sample paths of the MPS model. Since no actvations are used in this model, so all the paths are just linear functions. Here we sampled four sample paths from the MPS model. The structure of our model is as follows: the number of tensor nodes is $5$, the bond dimension is $8$ and the kernel dimension is $2$.}
\label{fig: pure_mps_sample_path}
\end{figure}

In Fig. \ref{fig: sample_path_mps}, we sample several random paths from the pure MPS GP. In Fig. \ref{fig: sample_path_mps_vars}, we plot the sample paths family of pure MPS GP with different standard deviation $\sigma_{\mathbf{A}}$. We find that the slope of the sample path increases hugely as the standard deviation $\sigma_{\mathbf{A}}$ of the prior of the tensor nodes increases.

\begin{figure}
\begin{subfigure}{.48\textwidth}
\centering
\includegraphics[width=0.9\linewidth]{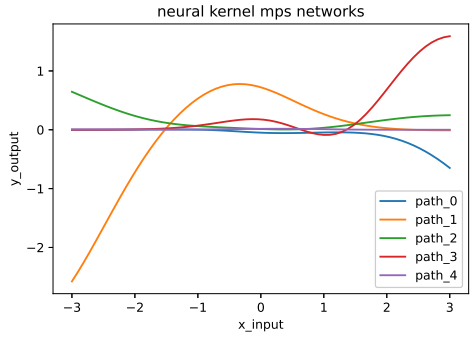}
\caption{plots of randomly sampled $4$ sample paths of neural kernel MPS.}
\label{fig: sample_path_nerual_kernel_mps}
\end{subfigure}
\begin{subfigure}{.48\textwidth}
\centering
\includegraphics[width=0.9\linewidth]{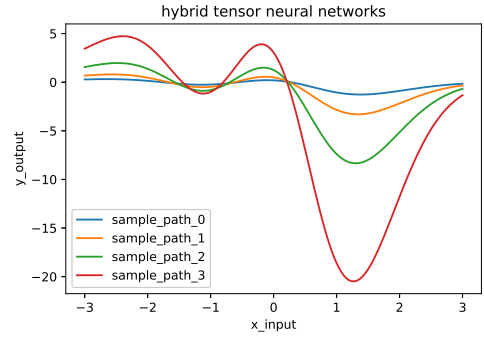}
\caption{plots of four sample path from the models with different variances.}
\label{fig: neural_kernel_mps_sample_path_vars}
\end{subfigure}
\begin{subfigure}{.48\textwidth}
\centering
\includegraphics[width=0.9\linewidth]{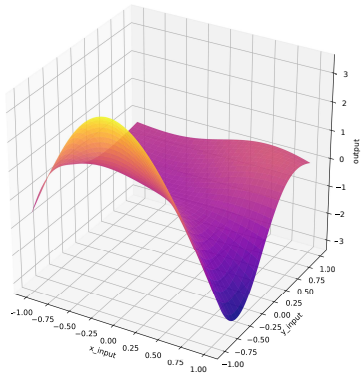}
\caption{plots of randomly sampled $4$ sample paths of tensor networks.}
\label{fig: neural_kernel_mps_2d_1}
\end{subfigure}
\begin{subfigure}{.48\textwidth}
\centering
\includegraphics[width=0.9\linewidth]{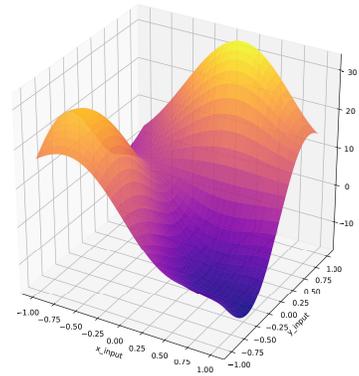}
\caption{plot of four sample path from the models with different variances.}
\label{fig: neural_kernel_mps_2d_2}
\end{subfigure}
\caption{In above figure, we plot the sample paths of the neural kernel MPS model. In this model, we introduce non-linearity by adding hidden neural layer and the activations.}
\label{fig: neural_kernel_mps}
\end{figure}

\begin{figure}
\begin{subfigure}{.48\textwidth}
\centering
\includegraphics[width=0.9\linewidth]{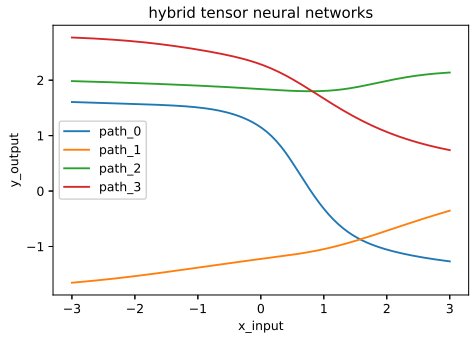}
\caption{plots of randomly sampled $4$ sample paths of MPS hidden neural network with $\sigma=0.5$.}
\label{fig: sample_paths_mps_hidden_neural_network_1}
\end{subfigure}
\begin{subfigure}{.48\textwidth}
\centering
\includegraphics[width=0.9\linewidth]{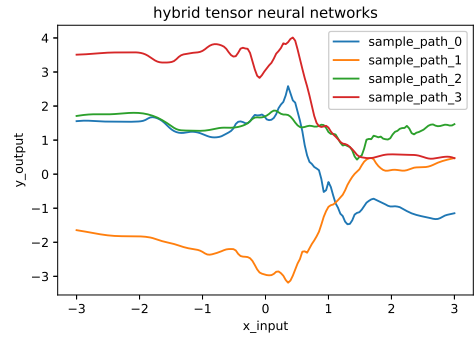}
\caption{plots of four sample paths from the models with variance $\sigma=1$.}
\label{fig: sample_paths_mps_hidden_neural_network_2}
\end{subfigure}\\
\begin{subfigure}{.48\textwidth}
\centering
\includegraphics[width=0.9\linewidth]{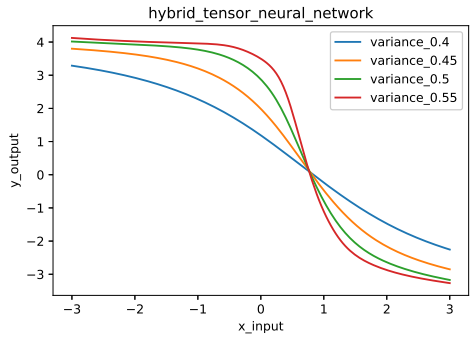}
\caption{plots of a randomly sampled family of sample paths of MPS hidden neural network with varying standard variation $\sigma$.}
\label{fig: sample_path_mps_hidden_neural_network_vars_1}
\end{subfigure}
\begin{subfigure}{.48\textwidth}
\centering
\includegraphics[width=0.9\linewidth]{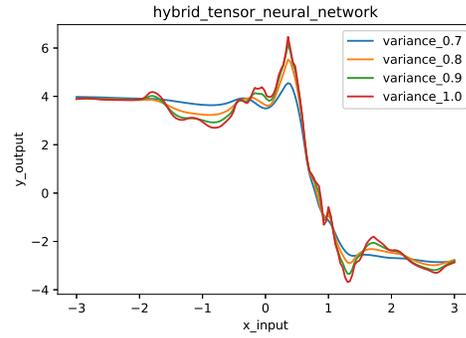}
\caption{plots of a family of sampled paths from MPS hidden layer neural network with different variances.}
\label{fig: sample_path_mps_hidden_neural_network_vars_2}
\end{subfigure}
\caption{In above figure, we plot the sample paths of the MPS hidden neural network.}
\label{fig: mps_hidden_neural_network}
\end{figure}

\begin{figure}
\begin{subfigure}{.48\textwidth}
\centering
\includegraphics[width=0.9\linewidth]{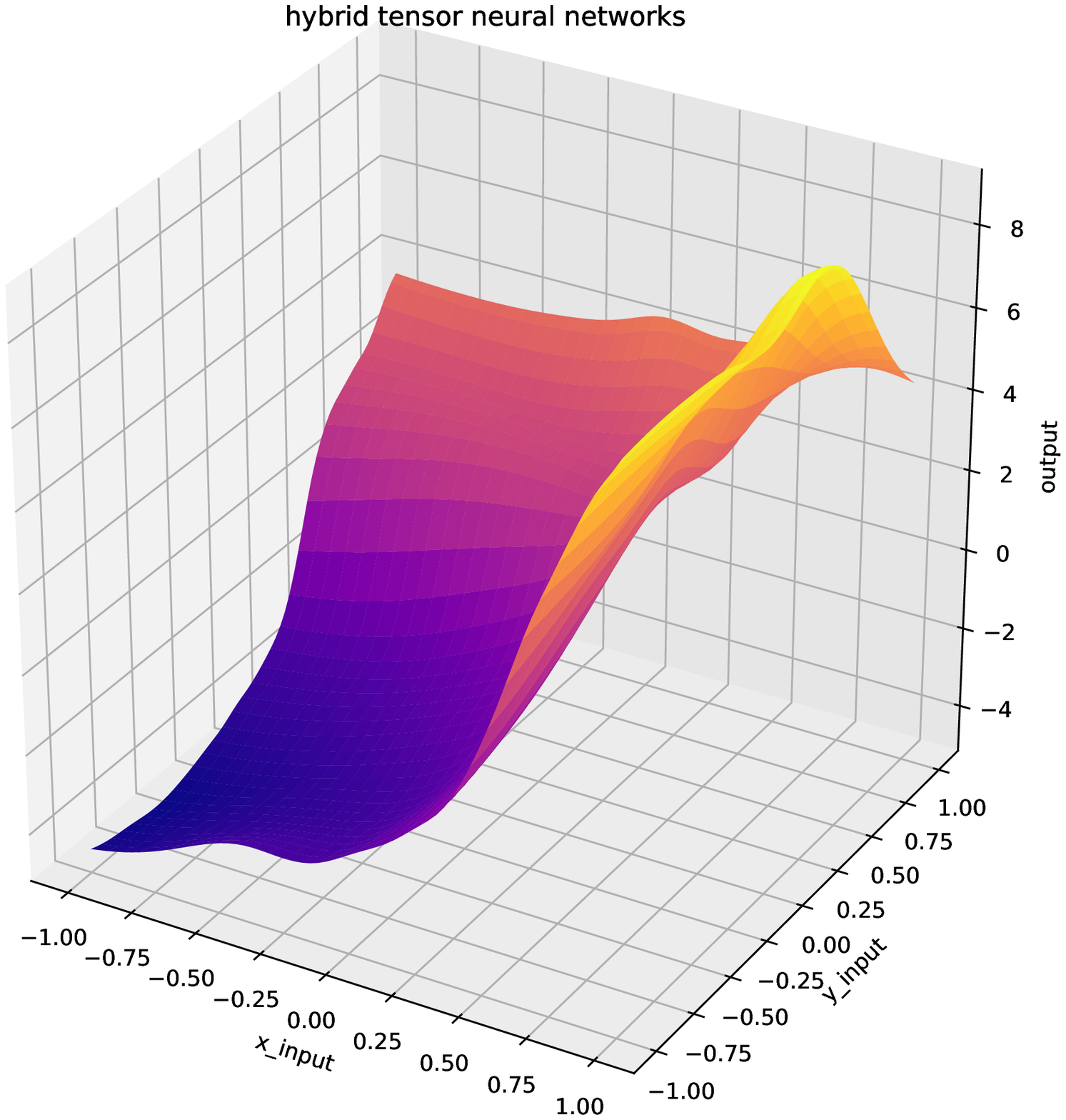}
\caption{sample path (surface) from MPS hidden neural network with $\sigma=0.7$.}
\label{fig: 2d_sample_path_mps_hidden_neural_network_1}
\end{subfigure}
\begin{subfigure}{.48\textwidth}
\centering
\includegraphics[width=0.9\linewidth]{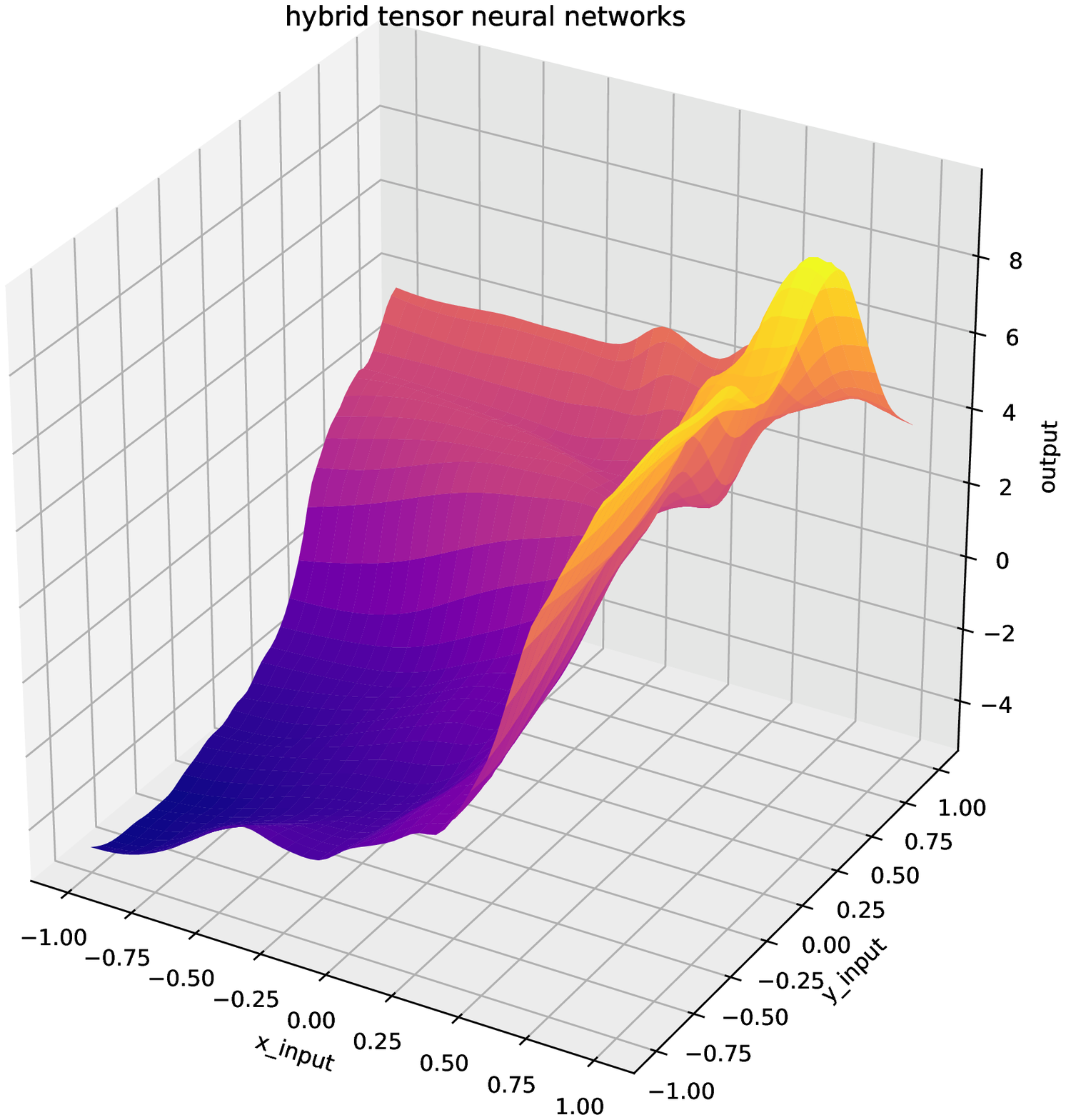}
\caption{plot of sampled path (surface) of MPS hidden neural network with 2 dimensional data vector.}
\label{fig: 2d_sample_path_mps_hidden_neural_network_2}
\end{subfigure}
\caption{In above figure, we plot the sample paths of the MPS hidden neural network model when the input data point is two dimensional vector.}
\label{fig: 2d_mps_hidden_nerual_network}
\end{figure}

In Fig. \ref{fig: neural_kernel_mps}, we plot the numerical experiment of the neural kernel MPS. In Fig. \ref{fig: sample_path_nerual_kernel_mps}, we plot the sample paths sampled from the neural kernel MPS. Since the introduction of neural hidden layer and the non-linear activation, the sample paths are not linear and more complicated than the sample paths by pure MPS model. In Fig. \ref{fig: neural_kernel_mps_sample_path_vars}, we plot one sample path varying as we increase the standard deviation $\sigma_{\mathbf{A}}$. As expected, the variation of the sample path increases as we increase the standard deviation $\sigma_{\mathbf{A}}$ of the tensor nodes prior. In the $d$ dimensional data vector case, the sample path is $d-1$ dimensional surface. In Fig. \ref{fig: neural_kernel_mps_2d_1} and \ref{fig: neural_kernel_mps_2d_2}, we plot the sample path of the neural kernel MPS model with two dimensional data points. Now the sample path is two dimensional surface.

In Fig. \ref{fig: sample_paths_mps_hidden_neural_network_1} and Fig. \ref{fig: sample_paths_mps_hidden_neural_network_2}, we compare the sample paths from MPS hidden neural network with increased standard deviation from $\sigma=0.5$ to $\sigma=1.0$. We find that as we increase the standard deviation of the prior, the sample paths become more complicated which captures greater data fitting ability. In Fig. \ref{fig: sample_path_mps_hidden_neural_network_vars_1} and Fig. \ref{fig: sample_path_mps_hidden_neural_network_vars_2}, the standard deviation $\sigma$ is increased and model complexity increases at accordingly.

In Fig. \ref{fig: 2d_sample_path_mps_hidden_neural_network_1} and Fig. \ref{fig: 2d_sample_path_mps_hidden_neural_network_2}, we plot the sample surfaces from the MPS hidden neural network in the two dimensional input data case. We can find that the complexity of the configures of two dimensional sample surface increases as the standard deviation increases. 

%!TEX root = main.tex

\section{Conclusion}
We study the infinite-width limit of tensor networks, acutally the pure MPS and the extended hybrid neural tensor networks: combining the MPS layer with neural layers. We prove that the pure MPS and the hybrid tensor networks converge to GP as the width of tensor network goes to infinity. For the pure MPS, since no non-linearity is injected into the model, the equivalent GP induced by the pure MPS is trivial in which each gaussian random variable indexed by a specific data point has zero uncertainty band. However, we can extend the pure MPS to more complicated structures by introducing hidden layers or using neural kernels. So we study the infinite limit of two kinds of hybrid tensor neural networks. We proved that these two structures converge to GP as the their widths go to infinity and calculate the equivalent GP of the infinite-width tensor network. In the numerical experiment, we explore the sample path and the properties of hyper-parameters of the tensor network induced GP and plot the sample path families by varying the standard deviations of the prior. The tensor networks have rich structures (intuitively the exponentially increasing contraction ways as the number of tensor nodes increases) and powerful fitting ability, so the function space supported by tensor networks has profound structures which allows well behaved limit processes. From the application perspective, over-parameterization and huge function space supported by tensor networks also lead to training instability, poor generalization and scaling difficulties which are essential problems in applying tensor network to big data set analysis.

%!TEX root = main.tex

\section*{Acknowledgements}
The authors wish to thank David Helmbold, Hongyun Wang, Qi Gong, Torsten Ehrhardt, Hai Lin and Francois Monard for their helpful discussions. Erdong Guo is grateful for the financial support by Ming-Ren Teahouse and Uncertainty Quantification LLC for this work.

\bibliographystyle{./ims.bst}
\bibliography{reference.bib}
\clearpage
%!TEX root = main.tex

\section*{A. Infinite Width Neural Networks}
\label{app:infinite_width_neural_net}
% Note: in this sample, the section number is hard-coded in. Following
% proper LaTeX conventions, it should properly be coded as a reference:

%In this appendix we prove the following theorem from
%Section~\ref{sec:textree-generalization}:
In this appendix we prove the equivalence of the infinite width one hidden layer Neural Net and GP.

\noindent
{\bf Lemma 1}
{\it For a sequence of random variable $\{X_{i}, i\in\{1, \cdots, k\}\}$,
\begin{align*}
\forall i\in \{1, \cdots, k\}, \quad X_{i} \sim \mathcal{N}(\mu_{i}, \sigma_{i}).
\end{align*}
If $\,\forall \mathbf{t} \in \mathrm{R}^{k}, \mathbf{t}\cdot\mathbf{X}\sim \mathcal{N}(\mu, \sigma)$, 
then 
\begin{align*}
    \mathbf{X}\sim\mathcal{N}(\mathbf{\mu}, \mathbf{\Sigma}).
\end{align*}
\noindent
{\bf Proof}. Since $\forall \mathbf{t}\in \mathrm{R}^{k}, \mathbf{t}\cdot\mathbf{X}\sim \mathcal{N}(\mu, \sigma)$, we have 
\begin{align*}
    &\mathrm{E}[\mathbf{t}\cdot\mathbf{X}] = \mathbf{t}\cdot\mathbf{\mu},\\
    &\mathrm{Var}[\mathbf{t}\cdot\mathbf{X}] = \mathbf{t}^{T}\cdot\mathbf{\Sigma}\cdot\mathbf{t}.
\end{align*}
We can write down the characteristic function $\phi_{\mathbf{X}}(t)$ of the random vector $\mathbf{X}=[x_{1}, \cdots x_{k}]$ as 
\begin{align*}
    \phi_{\mathbf{X}}(t) = \phi_{\mathbf{t}\cdot\mathbf{X}}(1) = \exp{(i\mathbf{t}\cdot\mathbf{\mu}-\frac{1}{2}\mathbf{t}^{T}\cdot\mathbf{\Sigma}\cdot\mathbf{t})},
\end{align*}
which is just the characteristic function of multi-variate normal distribution $\mathcal{N}(\mathbf{\mu}, \mathbf{\Sigma})$.
\hfill\BlackBox\\
}\\
\noindent
{\bf Theorem 1} 
{\it Let the response $f(\mathbf{x})$ of the one hidden layer neural network with $j$ neurons as 
\begin{align*}
    f(\mathbf{x}) = b + \sum_{j}w^{[1]}_{j}z_{j}(\mathbf{w}^{[0]}; \mathbf{x}),
\end{align*}
where
\begin{align*}
   &w_{j}^{[i]} \overset{\text{i.i.d.}}{\sim} \mathcal{N}(0, \sigma_{i}),\\
   &b \sim \mathcal{N}(0, \sigma_{b}).
\end{align*}
As $j$ goes to infinity, $f$ converge to the Gaussian Process,
\begin{align*}
    f\sim \mathcal{GP}(m, \Sigma),
\end{align*}
where $m$, and $\Sigma$ are the mean and covariance function. 
} 

\noindent
{\bf Proof}. 
Since all components of the neural layer weights $\mathbf{w}^{[0]}$ are i.i.d., we know that all the components of the output of the first layer network $z_{j}(\mathbf{w}^{[0]}; \mathbf{x})$ are i.i.d., \begin{align*}
   (z_{j}|\mathbf{w}^{[0]}, \mathbf{x})\overset{\text{i.i.d.}}{\sim} p(z_{j}),
\end{align*}
then we can easily verify that 
\begin{align*}
   (w^{[1]}_{j}z_{j}|\mathbf{w}^{[0]}, \mathbf{w}^{[1]}, \mathbf{x})\overset{\text{i.i.d.}}{\sim} p(w^{[1]}_{j}z_{j}),
\end{align*}
In neural networks, the activation functions are mostly well behaved, so the mean and the variance of the output $z_{j}$ are bounded, 
\begin{align*}
   &\mathrm{E}[w_{j}^{[1]}z_{j}] = 0,\\
   &\mathrm{Var}[w^{[1]}_{j}z_{j}] = \mathrm{Var}[w^{[1]}_{j}]\mathrm{Var}[z_{j}] < +\infty.
\end{align*}
By the central limit theorem, we know that as $j$ goes to $\infty$, the infinite sum $\sum_{j}w_{j}^{[1]}z_{j}$ will converge to a random variable with gaussian distribution, namely
\begin{align*}
    \frac{1}{n}\sum_{j}^{n}w_{j}^{[1]}z_{j} \sim \mathcal{N}(0, \sigma_{h}),
\end{align*}
where
\begin{align*}
    \sigma_{h}^{2} = \frac{1}{n}\sigma_{w^{[1]}}^{2}\sigma^{2}_{z}.
\end{align*}
Since random variable $b$ belongs to the gaussian distribution, then the sum of two gaussian variables $f(\mathbf{x})$, namely
\begin{align*}
    f(\mathbf{x}) = b + \sum_{j}w_{j}^{[1]}z_{j}(\mathbf{w}^{[0]}; \mathbf{x})\sim \mathcal{N}(0, \sqrt{\sigma_{b}^{2} + n\sigma_{h}^{2}})
\end{align*}
We can set $n\sigma^{2}_{w^{[1]}}$ to be fixed constant during the limit process to keep $f(\mathbf{x})$ well behaved. Now we have already proved that for a finite index $\mathbf{T}$ set which is just the training set $\{\mathbf{x}^{(i)}, i\in\mathrm{N}\}$, there exists a sequence of gaussian random variables $\{f(\mathbf{x}^{(i)}), \mathbf{x}^{(i)}\in\mathbf{T}\}$. To extend the finite sequence to the infinite index sequence such as the interval $\mathbf{T}$, we need to verify that the finite dimensional distributions satisfy the Kolmogorov consistency (extension) theorem. However, it is known that as long as the f.d.d.s. is a consistent distribution, then the stochastic process exists. In our neural network case, the f.d.d.s. also satisfies the two consistent conditions, then we proved that infinite width one hidden layer neural networks is equivalent to the gaussian process.
\hfill\BlackBox\\

\section*{B. Infinite Width Tensor Networks}
In this appendix, we will prove the theorems on the equivalence of the infinite width tensor network and GP. The main idea of the proofs are similar to the proof of Theorem \ref{the: neural_net_eq_gp}. 
\subsection*{B.1 Proof of Theorem \ref{the: pure_mps_eq_gp}}
\label{app: pure_mps_eq_gp_proof}
There are three indices in each tensor node $\mathbf{A}^{s_{i}}_{\alpha_{i}\alpha_{i+1}}$, so at first step we can contract the kernel $\Phi^{s_{1}, \cdots, s_{n}}(\mathbf{x})$ with the indices sequence $\{s_{1}, \cdots s_{n}\}$ in tensor network, and then we get 
\begin{align}
    \sum_{\{\alpha_{i}\}}Z_{\alpha_{1}\alpha_{2}}\cdots Z_{\alpha_{n}\alpha_{1}}, 
\end{align}
where
\begin{align}
    Z_{\alpha_{i}\alpha_{i+1}} = \sum_{s_{i}}A^{s_{i}}_{\alpha_{i}\alpha_{i+1}}\phi^{s_{i}}(\mathbf{x}_{i}).
\end{align}
It is easy to verify that all the components of the $Z_{\alpha_{i}\alpha_{i+1}}$ are i.i.d. since all components of one specific tensor node $\mathbf{A}^{s_{i}}_{\alpha_{i}\alpha_{i+1}}$ are i.i.d. We note here that in our assumption of Theorem \ref{the: pure_mps_eq_gp}, we just assume that all tensor nodes are independent, but not necessary to be i.i.d. For two specific implementations of the bond indices, namely two different sequences of value of the indices, $\{i_{1}, \cdots, i_{n}\}$ and $\{j_{1}, \cdots, j_{n}\}$, the product of each chain of the nodes are i.i.d.. We introduce a new notation for the production of the chain, 
\begin{align}
   K_{\{i_{1}\cdots i_{n}\}} = Z_{i_{1}i_{2}}\cdots Z_{i_{n}i_{1}}.
\end{align}
We know all the different possible evaluation of the indices sequence $\{\alpha_{1}\cdots\alpha_{n}\}$ of $K$ are i.i.d. As the width of the tensor network goes to infinity, the number of all the possible implementation of the indices sequence goes to infinity. By the C.L.T, $\psi$ converges to a GP.
\hfill\BlackBox\\

\subsection*{B.2 Proof of Theorem \ref{the: neural_kernel_mps_eq_gp}}
\label{app: neural_kernel_mps_eq_gp_proof}
Based on the work of Theorem \ref{the: pure_mps_eq_gp} proof, to prove Theorem \ref{the: neural_kernel_mps_eq_gp}, we just need to show that all the components of the neural kernel are independent. It is obvious that all component of the output of the neural kernel $a_{i}(\mathbf{w};\mathbf{x})$ are independent. By reshaping the output of the neural kernel $f^{l}$, we get the kernel $\Phi(\mathbf{x})$ and all the components are independent. Then by setting the width of the MPS to be infinite, using C.L.T and Theorem \ref{the: neural_kernel_mps_eq_gp}, the response of the neural kernel mps $\psi$ will converge to GP.  
\hfill\BlackBox\\

\subsection*{B.3 Proof of Theorem \ref{the: mps_hidden_neural_net_eq_gp}}
\label{app: mps_hidden_neural_net_eq_gp_proof}
By taking the infinite width limit of the hidden layer MPS, the response of the MPS $\psi^{l}$ are all i.i.d. which follow the normal distribution by Theorem \ref{the: pure_mps_eq_gp}, then it is easy to verify that $\sum_{l}w_{l}a(\psi^{l})$ follows gaussian distribution by C.L.T. and then $b + \sum_{l}w_{l}a(\psi^{l})$ is normal random variable. So $f$ converge to the GP as $l$ goes to infinity.
\hfill\BlackBox\\
%In this section, we will analyse the f.d.d.s. of the infinite width neural network. We will view this model as a non-parametric model so no weights will be updated. According to the results in above section, we know that the marginal distribution of every random variable $f(\mathbf{x}^{(i)})$ belongs to the gaussian distribution, 
%\begin{align}
%    f(\mathbf{x}^{(i)})\sim \mathcal{N}(0, \sigma_{f}).
%\end{align}
%Since the weights of the model are fixed and will not updated, then we know that the correlation between any two of the sequence of the random variables is $1$, namely
%\begin{align}
%    \text{corr}(f(\mathbf{x}^{(i)}), f(\mathbf{x}^{(j)})) = 1.
%\end{align}
%
%By above condition, we can write down the covariance matrix $\text{Cov}[f(\mathbf{x^{(i)})}, f(\mathbf{x}^{(j)})]$ as 
%\begin{align}
%   \text{Cov}[f(\mathbf{x}^{(i)}), f(\mathbf{x}^{(j)})] = \sigma_{f_{i}}\sigma_{f_{j}}.
%\end{align}
%And the joint distribution $p(f(\mathbf{x}^{(1)}), \cdots, f(\mathbf{x}^{(n)}))$ is 
%\begin{align}
%p(f(\mathbf{x}^{(1)}), \cdots, f(\mathbf{x}^{(n)})) = \mathcal{N}(\mathbf{0},K),    
%\end{align}
%where
%\begin{align*}
%    K = \text{Cov}[f(\mathbf{x}^{(i)}), f(\mathbf{x}^{(j)})].
%\end{align*}
%We can easily verify that 
%\begin{align}
%    p(f(\mathbf{x}^{(i)})|f(\mathbf{x}^{(j)})) = \lim_{\sigma_{f}\to 0}\mathcal{N}(\frac{\sigma_{i}}{\sigma_{j}}f(\mathbf{x}^{(j)}), \sigma_{f}) = \delta(\frac{\sigma_{i}}{\sigma_{j}}f(\mathbf{x}^{(j)})).
%\end{align}
\section*{C. Approximation of Covariance Function}
\label{app: cov_matrix_approx}
In this appendix, we will explain the calculation of the approximation of the covariance function of mps hidden neural network in detail. For a general function $f(\mathbf{x}; \mathbf{A})$, the expectation $\mathrm{E}_{A}[f(\mathbf{x}; \mathbf{A})]$ can be written as 
\begin{align}
    \mathrm{E}_{A}[f(\mathbf{x}; A)] &= \mathrm{E}_{A}[f(\mathbf{x}; A_{0}) + \nabla f(\mathbf{x}; A_{0})(A-A_{0}) + \frac{1}{2}(A-A_{0})^{T}\nabla^{2}f(\mathbf{x}; A_{0})(A-A_{0}) + R(\mathbf{x}; A)]\\
    & = f(\mathbf{x}; A_{0}) + \frac{1}{2}\mathrm{Var}[A]\cdot\mathrm{Diag}[H]+\mathrm{E}_{A}[R(\mathbf{x}, A)],
\end{align}
where
\begin{align*}
    R(\mathbf{x}, A) = \frac{1}{3!}f^{(3)}(\mathbf{x}; \xi)(A-A_{0})^{3},\quad \xi\in[A_{0}, \mathbf{x}].
\end{align*}

We write the second derivative matrix $H = \nabla^{2}f(\mathbf{x}; A_{0})$ as 
\begin{align}
    H = H^{T\frac{1}{2}}H^{\frac{1}{2}}
\end{align}
where $H^{\frac{1}{2}}$ is the formal square root of the the second derivative  matrix $H$. Since $H$ is real symmetric matrix, then $H^{\frac{1}{2}}$ exists and is symmetric.
\begin{align}
\mathrm{E}_{A}[\frac{1}{2}(A-A_{0})^{T}H(A-A_{0})] &= \frac{1}{2}\mathrm{Tr}\mathrm{E}_{A}[(A-A_{0})^{T}H^{T\frac{1}{2}}H^{\frac{1}{2}}(A-A_{0})]\\
&= \frac{1}{2}\mathrm{Tr}\mathrm{E}_{A}[(H^{\frac{1}{2}}(A-A_{0}))((A-A_{0})^{T}H^{T\frac{1}{2}}]\\
&=\frac{1}{2}\mathrm{Tr}[H^{\frac{1}{2}}\mathrm{E}_{A}[(A-A_{0})(A-A_{0})^{T}]H^{T\frac{1}{2}}]\\
&=\frac{1}{2}\mathrm{Tr}[\mathrm{Cov}[A-A_{0}, A-A_{0}]H]\\
&=\frac{1}{2}\mathrm{Tr}[\mathrm{Cov}[A, A]\cdot H]\\
&=\frac{1}{2}\mathrm{Var}[A]\cdot\mathrm{Diag}[H]
\end{align}
Here we use $\mathrm{Var}[A]$ to represent the diagonal part of the covariance matrix $Cov[A, A]$, and $\mathrm{Diag}[H]$ to represent the diagonal part of the second derivative matrix $\nabla^{2}f(\mathbf{x};A)$ of the integrated function $f(\mathbf{x}; A)$.
%\section*{Appendix B.}
%In this appendix, we show the detail calculation of the covariance function with Laplace approximation.
%
%The main idea of Laplace approximation is to use the Normal distribution located at the optimal point of the target distribution to substitute the target distribution in the integral. The integral with Gaussian measure is much more easy to be deal with and usually the analytical result can be obtained. 
%
%For the following integral 
%\begin{align*}
%\int{f(\mathbf{x},\theta)p(\theta)d\theta},
%\end{align*}
%where $f(\mathbf{x}, \theta)$ is arbitrary function of input data $\mathbf{x}$ and random variable $\theta$ (weights in model in our case), $p(\theta)$ is the probability distribution (prior distribution in our case) and we neglect the parameters here for convenience 
\hfill\BlackBox

\end{document}